\documentclass[a4paper]{article}

\usepackage[a4paper,top=3cm,bottom=2cm,left=3cm,right=3cm,marginparwidth=1.75cm]{geometry}

\usepackage{times}
\usepackage[utf8]{inputenc}
\usepackage{graphicx}
\usepackage{amsmath}
\usepackage{tabularx}
\usepackage{tabulary}
\usepackage[normalem]{ulem}
\usepackage{algorithm}
\usepackage{algpseudocode}

\title{ConvAMR: Abstract meaning representation parsing for legal document}
\author{Lai Dac Viet, Vu Trong Sinh, Nguyen Le Minh, Ken Satoh \\
\{vietld, sinhvtr, nguyenml\}@jaist.ac.jp, ksatoh@nii.ac.jp
}

\newcommand\tab[1][0.3cm]{\hspace*{#1}}

\begin{document}
\maketitle
\begin{abstract}
Convolutional neural networks (CNN) have recently achieved remarkable performance in a wide range of applications. In this research, we equip convolutional sequence-to-sequence (seq2seq) model with an efficient graph linearization technique for abstract meaning representation parsing. Our linearization method is better than the prior method at signaling the turn of graph traveling. Additionally, convolutional seq2seq model is more appropriate and considerably faster than the recurrent neural network models in this task. Our method outperforms previous methods by a large margin on both the standard dataset LDC2014T12. Our result indicates that future works still have a room for improving parsing model using graph linearization approach.

\end{abstract}

\section{Introduction}
Abstract Meaning Representation (AMR) forms a rooted acyclic directed graph that represents the content of a sentence. All nodes and edges of the AMR graph are labeled according to the sense of the words in a sentence. AMR parsing is the task of converting a given sentence to a corresponding graph. AMRs have been applied to several applications such as event extraction \cite{rao2017biomedical,garg2016extracting}, text summarization \cite{dohare2017text,liu2015toward} and text generation \cite{takase2016neural,songamr}. However, AMR annotation which requires a lot of human effort limits the outcome of data-driven approaches, one of which being neural network based methods \cite{neuralAMR,buys-blunsom:2017:SemEval}. Therefore, a highly accurate parser is necessary in order to intensify other applications which are based on AMR.

\begin{figure}
\includegraphics[width=\textwidth]{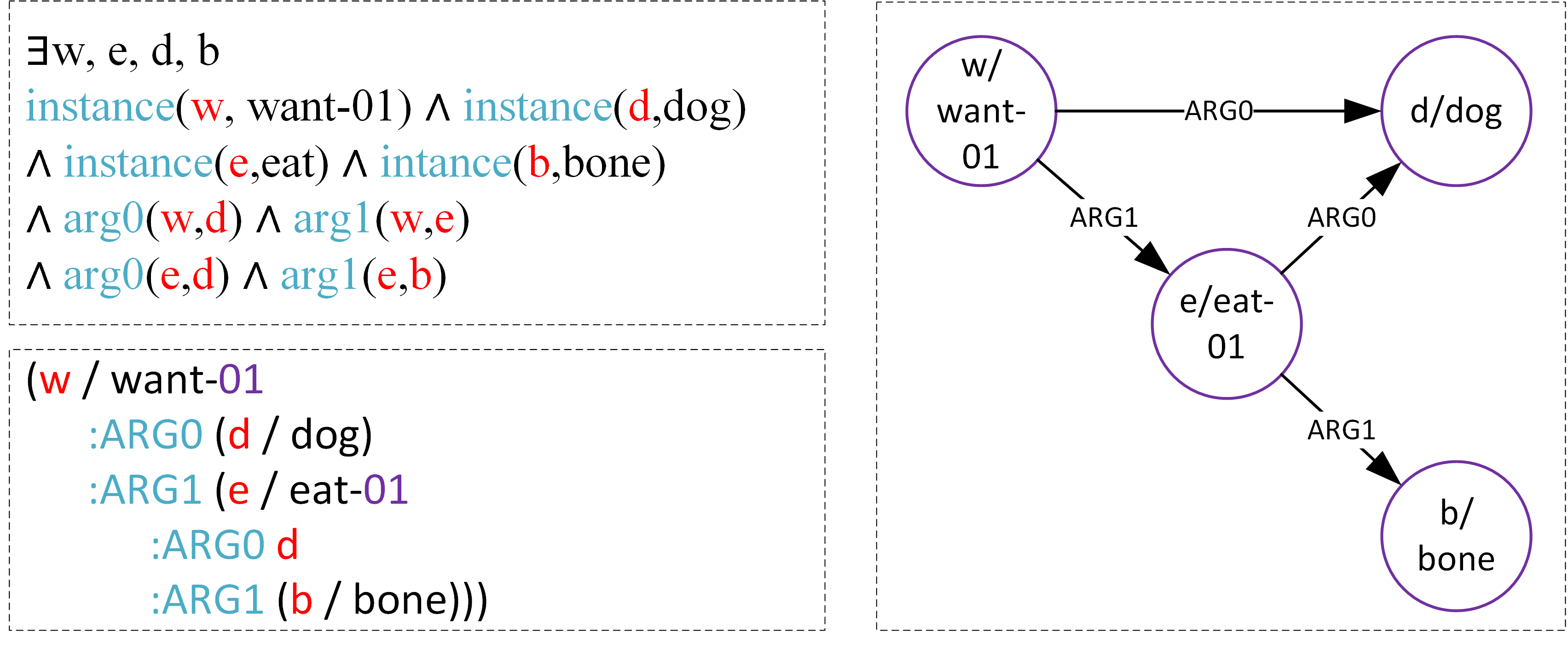}
\label{fig:format}
\caption{Three AMR formats of the same sentence \textit{"The dog wants to eat a bone"}. The conjunction form, the PENMAN notation and the graph are located on the top left, the bottom left and the right, respectively. }
\end{figure}

Three different ways are widely utilized to demonstrate AMR graphs. First, conjunction form represents AMR to measure the similarity between two AMR graphs and some logic applications. Secondly, the PENMAN notation is used on several occasions that are related to human reading and writing such as annotation and data observation. Thirdly, computer programs commonly store AMRs as graph structure in memory. Figure \ref{fig:format} illustrates three typical representation approaches. In an AMR graph, each node is managed using an unique ID called \textbf{variables}. The content of a node is expressed by a semantic \textbf{concept}, which can be an English word (e.g. $dog$) or a PropBank frameset (e.g. \textit{want-01}) or a special keyword (e.g. the "-" sign). The edge between two vertices is labeled using more than 100 relations including semantic relations (e.g. \textit{:location, :name}), and frameset argument index (e.g.\textit{:ARG0, :ARG1}). AMR also provides the inverse form of relations (e.g. \textit{:location} vs \textit{:location-of}).

To compare two semantic graphs, Cai et al \cite{cai2013smatch} introduced the \textbf{SMATCH} score. This score measures the level of structural overlapping between two structures. SMATCH score has been widely applied in measuring the accuracy of AMR parser.

\begin{figure}
\centering
\includegraphics[width=0.8\textwidth]{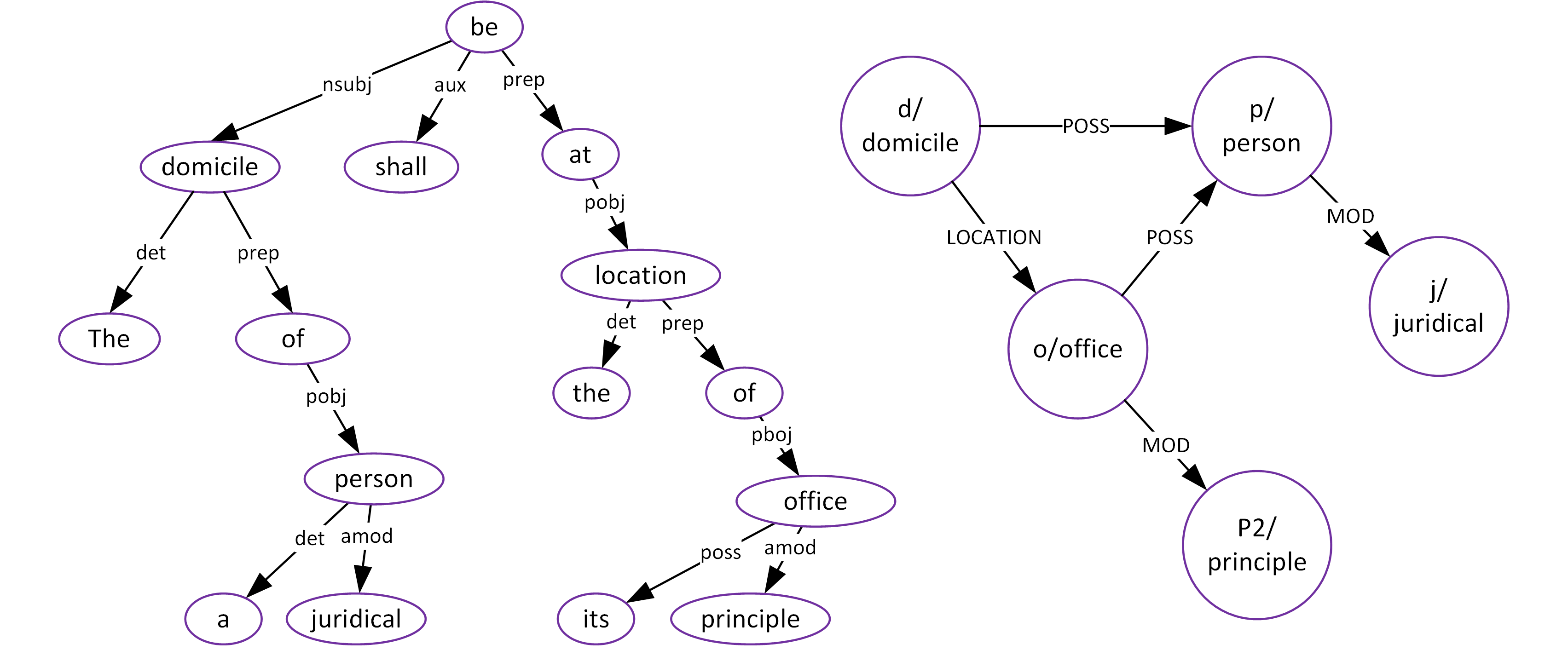}
\caption{Dependency tree and AMR graph}
\label{fig:dep}
\end{figure}

Transition-based parsers have made notable achievements in graph parsing such as dependency tree \cite{chen2014fast}. Currently, AMR parsers are benefiting from the power of this approach. Motivated by the analogy between dependency tree and AMR graph, Wang et al. \cite{wang:2015:NAACL-HLT} proposed the first transition system for parsing AMR graph. Figure \ref{fig:dep} illustrates the dependency tree and the AMR graph of the sentence: \textit{"The domicile of a juridical person shall be at the location of its principal office"}. These two structures share some nodes (e.g.\textit{domicile, person, juridical}), and their node interrelations (e.g.\textit{person - juridical}).  Wang et al. defined a two-stage process for their system: (1)parsing a sentence into a dependency tree using existing parsers such as Stanford parser and Charniak parser; (2) converting the obtained tree into AMR graph by an eight-action transition system. Their later works have investigated a richer feature set including co-reference, semantic role labeling, word cluster \cite{wang:2015:ACL-IJCNLP}; rich name entity tag, and ISI verbalization list\cite{wang:2016:SemEval}.

NeuralAMR \cite{neuralAMR} has succeeded at both AMR parsing and sentence generation as the result of a bootstrapping training strategy on a 20-million-sentence unsupervised dataset. An efficient adaptation of machine translation to AMR parsing by Barzdins et al \cite{barzdins2016RIGA} indicates that character-based features are better than word-based features. Targeting at the sparsity of the AMR graph data, the vocabulary of AMR is limited to 2000 in the work of Peng et al \cite{peng2017addressing}. The work of Ballesteros et al has combined recurrent neural network and transition system into a deep transition model \cite{ballesteros2017amr}. Among those methods, the information is encoded in LSTM hidden state using embedding vector and syntactic features instead of gathering a large number of features which are introduced in the conventional transition method.

Although recent studies have utilized Long Short-Term Memory (LSTM) in AMR parsing \cite{neuralAMR,ballesteros2017amr}, there are several disadvantages of employing LSTM compared to CNN. First, LSTM models long dependency, which might be noise to generate a linearized graph, whereas CNN provides a shorter dependency which is advantageous to generate graph traversal. Secondly, LSTM requires a chronologically computing process that restrains the ability of parallelization; on the contrary, CNN enables simultaneous parsing. In this paper, we present the first success of applying convolutional seq2seq in AMR parsing. The main contributions of this research are:

\begin{itemize}
\item An outstanding performance with 5 points SMATCH score improvement resulted from the proposed AMR parsing model using depth-first-search graph linearization and convolutional seq2seq network.
\item A new public AMR test \footnote{https://github.com/nguyenlab/crest} set of legal document.
\item The first study of AMR parsing in the legal domain.
\end{itemize}

\section{Method}
In this section, we first present the formalization of the AMR parsing task. We then demonstrate in detail two main parts of our model: the graph conversion including linearization and de-linearization presented in section \ref{sec:linearization}, and convolutional seq2seq model presented in section \ref{sec:model}.

Given the training dataset $\langle S,G \rangle$ where $S$ and $G$ stand for the set of sentences and the set of corresponding AMR graphs, our supervised learning model with parameter set $\theta$ maximizes the following problem:
$$
\hat{\theta} = \operatorname*{argmax}_\theta \sum_{i}{ P(G_i| S_i, \theta)}
$$

\subsection{Graph linearization and de-linearization}
\label{sec:linearization}
Seq2seq model requires sequential representation of features and labels, therefore, the AMR graph must be presented as a sequence. However, the raw AMR text cannot be an appropriate format due to its imbalance of tokens. Raw AMR text contains too many round brackets and variables which present less semantic information than other components such as concepts, constants, and English words. Unlike the prior work \cite{neuralAMR}, in our model, the graphs pass through a much simpler pre-processing series which consists of variable removal, graph linearization, and infrequent word replacement. For stripping the AMR text, we modified the \textbf{depth-first-search traversal} from the work of Kontas et al \cite{neuralAMR} in the way of marking the end of a path. The left parentheses are ignored and the right parentheses are replaced by doubling the concept of the terminal node. 

The process of recovering the stripped text from the graph is called de-linearization. The graph which contains multiple nodes of a single concept might not be perfectly reversed because those nodes have been collapsed into one. We show the level of information loss corresponding to each dataset in section \ref{tbl:statistics}. Table \ref{tbl:conversion} demonstrates the converting process.

\begin{table}[h]
\caption{Graph conversion process in detail}
\label{tbl:conversion}
\centering
\begin{tabular}{| l | l |}
      \hline 
      Original article  & No abuse of rights is permitted. \\
      \hline
               &  (p / permit-01 :polarity - \\
      AMR Text & \tab:ARG1 (a / abuse-01 \\
               & \tab\tab:ARG1 (r / right-05))) \\
    \hline
                       &  (permit-01 :polarity - \\
       Variable removal& \tab :ARG1 (abuse-01 \\
                       & \tab\tab :ARG1 (right-05))) \\ 
    \hline 
      Linearization & permit-01 :polarity - - :ARG1 abuse-01 :ARG1 right-05 right-05 \\
                    & abuse-01 permit-01 \\
    \hline  
\end{tabular}
\end{table}

We conducted the measurement of information loss to prove the efficiency of this graph conversion method. All graphs in the official AMR corpus $G$ were passed through a full linearization process to get the linearized versions. These sequences were then the input of recovering process to obtain the AMR graph set $G'$. The information loss $L(G)$ is calculated by equation \ref{eq:loss} from the \textbf{SMATCH} score. The result of the test is presented in table \ref{tbl:statistics}.

\begin{equation}
\label{eq:loss}
L(G) = 1 - \frac{1}{n}\sum_{n}{Smatch(G_i, G'_i)}
\end{equation}

\subsection{Convolutional sequence to sequence model}
\label{sec:model}
Our proposal is to utilize three different seq2seq models which have showed their strengths in machine translation. They are the combination of a convolutional encoder and an LSTM decoder; and the fully convolutional seq2seq model. The first model uses a multilayer bi-directional LSTM encoder to produce hidden states from the input. The decoder gathers the hidden states and then generates output with attention mechanism. We made a further modification by supplementing a dropout layer locating between two consecutive LSTM layers. The second model bases on the work of Gehring et al \cite{gehring2016convolutional} where the bi-directional LSTM is alternated by a convolutional encoder. The final model fully applies convolutional neural network with attention mechanism \cite{gehring2017convolutional}.

\subsection{Data annotation}
The Semeval competitions allowed participants to access multiple AMR corpus annotated manually but no large corpus has been made accessible to the public. Especially, there is no open AMR resource for any specific domain such as the juristic document or scientific document. Therefore, we manually annotated a corpus for the English version of the Japan Civil Code. The code is organized in multiple levels including chapter, part, article, paragraph, and sentence. 

The pre-processing consists of the following steps: gathering articles, removing all article prefixes and article IDs, then splitting the article into sentences. We labeled each sentence with an ID containing the article name, the paragraph index, and the sentence index. To annotate the sentences, we used the web-based editor \footnote{https://amr.isi.edu/editor.html} provided by ISI group. This editor provides a combination of command line and graphical interface. The Propbank corpus is integrated into the search engine to minimize the time it takes to choose a proper meaning of the words. Two annotators are given a list of article sentences and annotate corpus independently. After finishing their own works, the annotators are invited to discuss and aggregate their outcomes into a single result. We call this dataset \textbf{JCivilCode-1.0}. The statistics of this corpus is presented in table \ref{tbl:statistics}.

\section{Experiment \& Result}
\label{sec:experiment}
We conducted the experiment on two datasets in different domains. The first one is the official dataset \textbf{LDC2014T12}, which we designed the first experiment configuration with. The second configuration was made on our self-annotated dataset by mixing the training set and the validation set of LDC2014T12 together with JCivilCode-1.0 as the test set. We decided to train and test on two different domains because the number of pair of sentences and graphs are not too large. The performance of the proposed approaches is assessed using SMATCH score. To compare our model with other ones, we collected the performance results of other works on LDC2014T12 from the original paper. We also run their best public pre-trained model on JCivilCode-1.0.

\begin{table}[h]
\caption{Dataset characteristics.}
\label{tbl:statistics}
\centering  
  \begin{tabular}{l|c|c}
     & \textbf{LDC2014T12} & \textbf{JCivilCode-1.0} \\
     \hline
    Train set & 10,312 & 0\\  
    Valid set & 1,368 & 0 \\
    Test set & 1,371 & 157 \\
    Domain & News & Legal \\
    Information loss & 0.21 & 0.20 \\
\end{tabular}
\end{table}

Table \ref{tbl:result} shows that our proposed model outperformed both the transition-based methods and the neural-based methods on LDC2014T12 whereas our methods are much simpler than prior works. The NeuralAMR lies on an intensive preprocessing with graph simplification and strong name-entity anonymization. The stack-LSTM model gathers many types of syntactic features including name entity, part-of-speech, dependency tree. Moreover, CAMR relies on rich features of a single node, node pair, path, distance, action \cite{wang:2015:NAACL-HLT} and semantic role labeling \cite{wang:2015:ACL-IJCNLP}. On the other hand, our models employ only word embedding as feature after a three-step preprocessing as described in section \ref{sec:linearization}. 

Linearization method might create two foreseeable issues though it significantly increased the accuracy of neural network method. First, entity redundancy occurs if the graph contains multiple nodes who share an identical concept. The second issue is the syntax error of the output because the neural network does not guarantee that the output follows the PENMAN notation. Table \ref{tbl:sample} shows some sample of JCivilCode-1.0 and output of our model. The bold words in the table show the error that our model generated.

\begin{table}[!h]
\caption{Dataset information}
\label{tbl:result}
\centering  
  \begin{tabular}{l|c|c}
      \textbf{Method} & \textbf{LDC2014T12} & \textbf{JCivilCode-1.0} \\
    \hline
      NeuralAMR \cite{neuralAMR} & 0.62 & - \\
      Stack LSTM \cite{ballesteros2017amr} & 0.64 & - \\
      CAMR \cite{wang:2015:ACL-IJCNLP} & 0.66 & - \\
      CAMR \cite{wang:2016:SemEval}& 0.66 & 0.46 \\
    \hline
      Conv encoder, LSTM decoder & 0.69 & 0.59 \\
      Fully conv seq2seq & \textbf{0.71} & \textbf{0.60} \\
\end{tabular}
\end{table}

\begin{table}[h]
\caption{Two type of structural error.}
\label{tbl:sample}
\centering  
  \begin{tabular}{|p{6cm}|p{6cm}|}
  \hline
      \textbf{Gold standard} & \textbf{System} \\
    \hline
    \multicolumn{2}{|p{12cm}|}{\textbf{[Node collision]} A person who has become subject to the ruling of commencement of guardianship shall be an adult ward, and a guardian of an adult shall be appointed for him/her. } \\
    \hline
      (a / and \newline
      \tab:op1 (a2 / adult \newline
      \tab\tab :domain \textbf{(p / person} \newline
      \tab\tab\tab :ARG1-of (s / subject-01 \newline
      \tab\tab\tab\tab :ARG2 (c / commence-01 \newline
      \tab\tab\tab\tab\tab :ARG1 (g / guard-01))))) \newline
      \tab :op2 (a3 / appoint-01 \newline
      \tab\tab :ARG1 \textbf{(p2 / person)} \newline
      \tab\tab :ARG2 (g2 / guardian \newline
      \tab\tab\tab :poss p)))
    &
      (a0 / and \newline
      \tab :op1 (x0 / <<unk>> \newline
      \tab\tab :domain \textbf{(p0 / person} \newline
      \tab\tab\tab :ARG1-of (s0 / subject-01 \newline
      \tab\tab\tab\tab :ARG2 (c0 / commence-01 \newline
      \tab\tab\tab\tab\tab :ARG1 (g0 / guard-01))))) \newline
      \tab :op2 (a1 / appoint-01 \newline
      \tab\tab :ARG1 \textbf{p0} \newline
      \tab\tab\tab :ARG2 x0 \newline
      \tab\tab\tab\tab :poss p0))      \\
    \hline
      \multicolumn{2}{|p{12cm}|}{\textbf{[Syntax error]} Unless otherwise provided by applicable laws, regulations or treaties, foreign nationals shall enjoy private rights. } \\
    \hline
      (e / enjoy-01 \newline
      \tab :ARG0 (n / national \newline
      \tab\tab :mod (f / foreign)) \newline
      \tab :ARG1 (r / right-05 \newline
      \tab\tab :ARG1-of (p / private-02)) \newline
      \tab :condition (p2 / provide-01 :polarity - \newline
      \tab\tab\tab :OR (o / or \newline
      \tab\tab\tab\tab :op1 (l / law \newline
      \tab\tab\tab\tab\tab :mod (a / applicable)) \newline
      \tab\tab\tab\tab :op2 (r2 / regulate-01) \newline
      \tab\tab\tab\tab :op3 (t / treaty)))) 
    &
      (e0 / enjoy-01 \newline
      \tab :ARG0 (n0 / national\newline
      \tab\tab :mod (f0 / foreign))\newline
      \tab :ARG1 (x0 / <<unk>>)\newline
      \tab :ARG1-of x0\newline
      \tab :condition (p0 / provide-01\newline
      \tab\tab :polarity (x1 / -)\newline
      \tab\tab \textbf{x0} \newline
      \tab\tab\tab \textbf{(o0 / or} \newline
      \tab\tab\tab\tab \textbf{:op1 (l0 / law} \newline
      \tab\tab\tab\tab\tab \textbf{:mod x0)} \\
    \hline
\end{tabular}
\end{table}

\section{Conclusion}
We published the first release of a testing set of Japan Civil Code for AMR parsing. We presented the efficiency of the convolutional seq2seq model on Abstract Meaning Representation parsing. By using a simple but effective graph linearization methods, our model achieved a competitive accuracy. The result indicates a certain possibility of higher performance on many application basing on AMR. However, this method revealed two technical issues that we plan to investigate more in future research. 

\section{Acknowledgement}
This work was supported by JST CREST Grant Number JPMJCR1513, Japan. The authors would like to thank our colleagues and reviewers for their intensive comments and suggestions.

\bibliography{main.bbl}

\begin{thebibliography}{10}
\providecommand{\url}[1]{\texttt{#1}}
\providecommand{\urlprefix}{URL }

\bibitem{ballesteros2017amr}
Ballesteros, M., Al-Onaizan, Y.: Amr parsing using stack-lstms. EMNLP  (2017)

\bibitem{barzdins2016RIGA}
Barzdins, G., Gosko, D.: {RIGA} at semeval-2016 task 8: Impact of smatch
  extensions and character-level neural translation on {AMR} parsing accuracy.
  CoRR  abs/1604.01278 (2016), \url{http://arxiv.org/abs/1604.01278}

\bibitem{buys-blunsom:2017:SemEval}
Buys, J., Blunsom, P.: Oxford at semeval-2017 task 9: Neural amr parsing with
  pointer-augmented attention. In: SemEval-2017. pp. 914--919 (August 2017)

\bibitem{cai2013smatch}
Cai, S., Knight, K.: Smatch: an evaluation metric for semantic feature
  structures. In: ACL (2). pp. 748--752 (2013)

\bibitem{chen2014fast}
Chen, D., Manning, C.: A fast and accurate dependency parser using neural
  networks pp. 740--750 (2014)

\bibitem{dohare2017text}
Dohare, S., Karnick, H.: Text summarization using abstract meaning
  representation. arXiv preprint arXiv:1706.01678  (2017)

\bibitem{garg2016extracting}
Garg, S., Galstyan, A., Hermjakob, U., Marcu, D.: Extracting biomolecular
  interactions using semantic parsing of biomedical text. In: AAAI. pp.
  2718--2726 (2016)

\bibitem{gehring2016convolutional}
Gehring, J., Auli, M., Grangier, D., Dauphin, Y.N.: A convolutional encoder
  model for neural machine translation. arXiv preprint arXiv:1611.02344  (2016)

\bibitem{gehring2017convolutional}
Gehring, J., Auli, M., Grangier, D., Yarats, D., Dauphin, Y.N.: Convolutional
  sequence to sequence learning. arXiv preprint arXiv:1705.03122  (2017)

\bibitem{neuralAMR}
Konstas, I., Iyer, S., Yatskar, M., Choi, Y., Zettlemoyer, L.: Neural {AMR:}
  sequence-to-sequence models for parsing and generation. CoRR  (2017)

\bibitem{liu2015toward}
Liu, F., Flanigan, J., Thomson, S., Sadeh, N., Smith, N.A.: Toward abstractive
  summarization using semantic representations. In: NAACL. pp. 1077--1086
  (2015)

\bibitem{peng2017addressing}
Peng, X., Wang, C., Gildea, D., Xue, N.: Addressing the data sparsity issue in
  neural amr parsing. EACL  (2017)

\bibitem{rao2017biomedical}
Rao, S., Marcu, D., Knight, K., Daum{\'e}~III, H.: Biomedical event extraction
  using abstract meaning representation. BioNLP 2017 pp. 126--135 (2017)

\bibitem{songamr}
Song, L., Zhang, Y., Peng, X., Wang, Z., Gildea, D.: Amr-to-text generation as
  a traveling salesman problem. In: EMNLP (2016)

\bibitem{takase2016neural}
Takase, S., Suzuki, J., Okazaki, N., Hirao, T., Nagata, M.: Neural headline
  generation on abstract meaning representation. In: EMNLP. pp. 1054--1059
  (2016)

\bibitem{wang:2016:SemEval}
Wang, C., Pradhan, S., Pan, X., Ji, H., Xue, N.: Camr at semeval-2016 task 8:
  An extended transition-based amr parser. In: SemEval-2016. pp. 1173--1178
  (June 2016)

\bibitem{wang:2015:ACL-IJCNLP}
Wang, C., Xue, N., Pradhan, S.: Boosting transition-based amr parsing with
  refined actions and auxiliary analyzers. In: ACL-IJCNLP (Vol 2). pp. 857--862
  (July 2015)

\bibitem{wang:2015:NAACL-HLT}
Wang, C., Xue, N., Pradhan, S.: A transition-based algorithm for amr parsing.
  In: NAACL:HLT. pp. 366--375 (May--June 2015)

\end{thebibliography}

\end{document}